\documentclass[letterpaper, 10 pt, conference]{ieeeconf}  % Comment this line out if you need a4paper

\IEEEoverridecommandlockouts                              % This command is only needed if 
\usepackage{setspace}
\usepackage{filecontents}
\usepackage{cite}
\usepackage{caption}
\usepackage{subfiles}
\usepackage{amsmath,amssymb,amsfonts}
\usepackage[short]{optidef}
\usepackage{algorithmic}
\usepackage{graphicx}
\usepackage{textcomp}
\usepackage{xcolor}
\usepackage{etex}
\usepackage{tikz}
\usetikzlibrary{matrix}
\usetikzlibrary{backgrounds}
\usetikzlibrary{calc}
\usetikzlibrary{shapes.misc}
\usetikzlibrary{decorations.pathmorphing,patterns}

\usepackage{pgfplots}
\usetikzlibrary{calc}
\usepackage{url}

\usepackage{hyperref}
\hypersetup{
	colorlinks,
	linkcolor={red!65!black},
	citecolor={green!65!black},
	urlcolor={blue!65!black}
}
\usepackage{tabularx}
\usepackage{hhline}
\usepackage{subcaption}
\newcommand{\bs}[1]{\boldsymbol{#1}}
\newcommand{\hT}{^\text{T}}

\title{\LARGE \bf
Linear-Time Contact and Friction Dynamics in Maximal Coordinates \\using Variational Integrators
}

\author{Jan Br\"udigam$^{1}$, Jana Janeva$^{1}$, Stefan Sosnowski$^{1}$, and Sandra Hirche$^{1}$% <-this % stops a space
\thanks{$^{1}$All authors are with the Chair of Information-oriented Control (ITR), Technical University of Munich, Munich, Germany
        {\tt\small \{jan.bruedigam, jana.janeva, sosnowski, hirche\}@tum.de}}%
}

\begin{document}

\maketitle
\thispagestyle{empty}
\pagestyle{empty}

%%%%%%%%%%%%%%%%%%%%%%%%%%%%%%%%%%%%%%%%%%%%%%%%%%%%%%%%%%%%%%%%%%%%%%%%%%%%%%%%
\begin{abstract}
Simulation of contact and friction dynamics is an important basis for control- and learning-based algorithms. However, the numerical difficulties of contact interactions pose a challenge for robust and efficient simulators. A maximal-coordinate representation of the dynamics enables efficient solving algorithms, but current methods in maximal coordinates require constraint stabilization schemes. 

Therefore, we propose an interior-point algorithm for the numerically robust treatment of rigid-body dynamics with contact interactions in maximal coordinates. Additionally, we discretize the dynamics with a variational integrator to prevent constraint drift. Our algorithm achieves linear-time complexity both in the number of contact points and the number of bodies, which is shown theoretically and demonstrated with an implementation. Furthermore, we simulate two robotic systems to highlight the applicability of the proposed algorithm.
\end{abstract}
%%%%%%%%%%%%%%%%%%%%%%%%%%%%%%%%%%%%%%%%%%%%%%%%%%%%%%%%%%%%%%%%%%%%%%%%%%%%%%%%

\section{Introduction}
Many modern robotics applications comprise systems that interact with the environment. 
Exemplary scenarios are bipedal or quadrupedal walking for inspection and load carrying tasks \cite{hutter_anymal_2016,kuindersma_optimization-based_2016}, or grasping and manipulation of objects \cite{kawasaki_dexterous_2002,butterfass_dlr-hand_2001}. As a result, state of the art control methods, such as nonlinear model-predictive control or learning-based methods, require fast and accurate simulations of such systems to quickly predict future states or generate training data.

Due to the numerical difficulties of rigid contact interactions with dry friction, a number of simulators with different approaches to handling contact and friction were developed, with overviews given in \cite{erez_simulation_2015,horak_similarities_2019}. Generally, the physically accurate description of contact and friction is discontinuous, leading to numerical instability and difficulties in calculating gradients when in contact. One approach to avoid numerical pitfalls is to formulate contact dynamics as soft contacts, for example as in MuJoCo \cite{todorov_mujoco_2012}. While this strategy offers reliable performance and allows for the computation of gradients, it comes at the cost of lacking physical accuracy. It has been shown that for soft models, data-driven methods require unreasonably high degrees of softness to make progress during learning \cite{parmar_fundamental_2021}.

Consequently, an approach for treating hard contacts while still enabling differentiation has recently been proposed \cite{cleach_linear_2021}. In this work, contact-implicit model-predictive control based on an interior-point method for simulating the dynamics is performed for a variety of systems. While this approach is promising, it does not achieve linear-time computational complexity for calculating contact and friction constraints due to a minimal-coordinate (also called generalized or joint coordinates) formulation. 

\begin{figure}[t] 
	\centering
	\resizebox{0.49\textwidth}{!}{\includegraphics{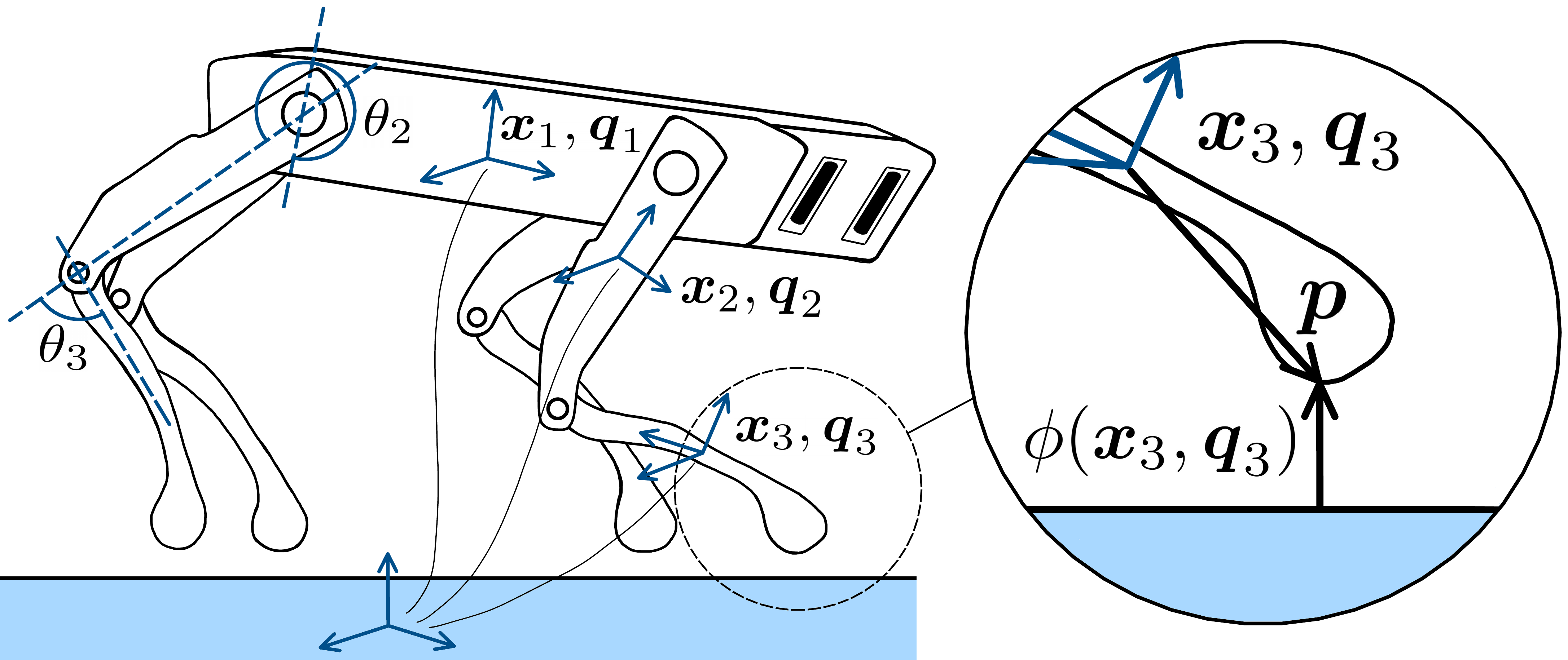}}
	\caption{A quadrupedal robot visualizing the difference between minimal coordinates (left leg), and maximal coordinates (right leg). On the right, the formulation of the signed distance function in maximal coordinates is shown.}\label{fig:contact_comp}
\end{figure}

Linear-time complexity for articulated mechanisms with contact described in maximal coordinates has been demonstrated \cite{weinstein_dynamic_2006}. As opposed to minimal coordinates, maximal coordinates parameterize each body in an articulated mechanism with its full six-dimensional configuration and enforce constraints, such as joints, with Lagrange multipliers \cite{baraff_linear-time_1996}. However, maximal-coordinate approaches, such as the method in \cite{weinstein_dynamic_2006}, typically require constraint stabilization schemes \cite{baumgarte_stabilization_1972} which introduce their own numerical difficulties. Figure \ref{fig:contact_comp} visualizes the difference between the two coordinate systems.

Variational (symplectic) integrators \cite{marsden_discrete_2001} have been shown to prevent constraint drift for maximal-coordinate dynamics with equality constraints while performing competitive to minimal coordinates \cite{brudigam_linear-time_2021}. However, this formulation has not been derived for systems with inequality constraints, such as those occurring in contact and friction scenarios. 

The main contribution of this paper is therefore a variational integrator for articulated systems with environmental contact and friction interactions in maximal coordinates. The computational complexity of treating contact and friction scales linearly both with the number of bodies in a mechanism, and the number of contact points. As opposed to other maximal-coordinate methods, all constraints of articulated mechanisms in our method are formulated on a position level, and we do not require any stabilization schemes.

The remainder of the paper is structured as follows: in Sec. \ref{sec:background} we briefly review the contact and friction model used in this paper and give a short overview of maximal coordinates in combination with variational integrators. Section \ref{sec:method} describes the formulation of contact and friction in maximal coordinates, shows the linear-time complexity, and proposes an interior-point algorithm for solving the dynamics and constraint equations. In Sec. \ref{sec:eval} we verify the linear-time complexity and give two simulation examples of robotic systems. Conclusions are drawn in Sec. \ref{sec:conclusions}.

\section{Background}\label{sec:background}
In this section, the description of contact and friction dynamics used for the remainder of the paper is reviewed, and the treatment of rigid body dynamics in maximal coordinates with variational integrators is described.

\subsection{Contact and Friction}
The dynamics of contact and friction for rigid bodies has been treated rigorously by several authors, see, for example, \cite{steward_implicit_1996} and \cite{anitescu_formulating_1997}, and we will summarize the main concepts. 

Rigid contact between a body with configuration $\bs{z}$ and the environment can be described with a signed distance function 
\begin{equation}
	\phi(\bs{z}) \geq 0,\label{eq:contact_constraint}
\end{equation}
where the body is in contact with the environment for $\phi(\bs{z}) = 0$. Figure \ref{fig:contact_comp} shows the signed distance for the foot of a quadrupedal robot. As mentioned above, due to the non-smoothness of this contact constraint, some numerical solvers allow for a small amount of penetration of the environment, i.e., soft contact, which improves the numerical stability at the cost of physical correctness. 
However, the method presented in this paper treats contact interactions as fully rigid.

\begin{figure}[b] 
	\centering
	\resizebox{0.49\textwidth}{!}{\includegraphics{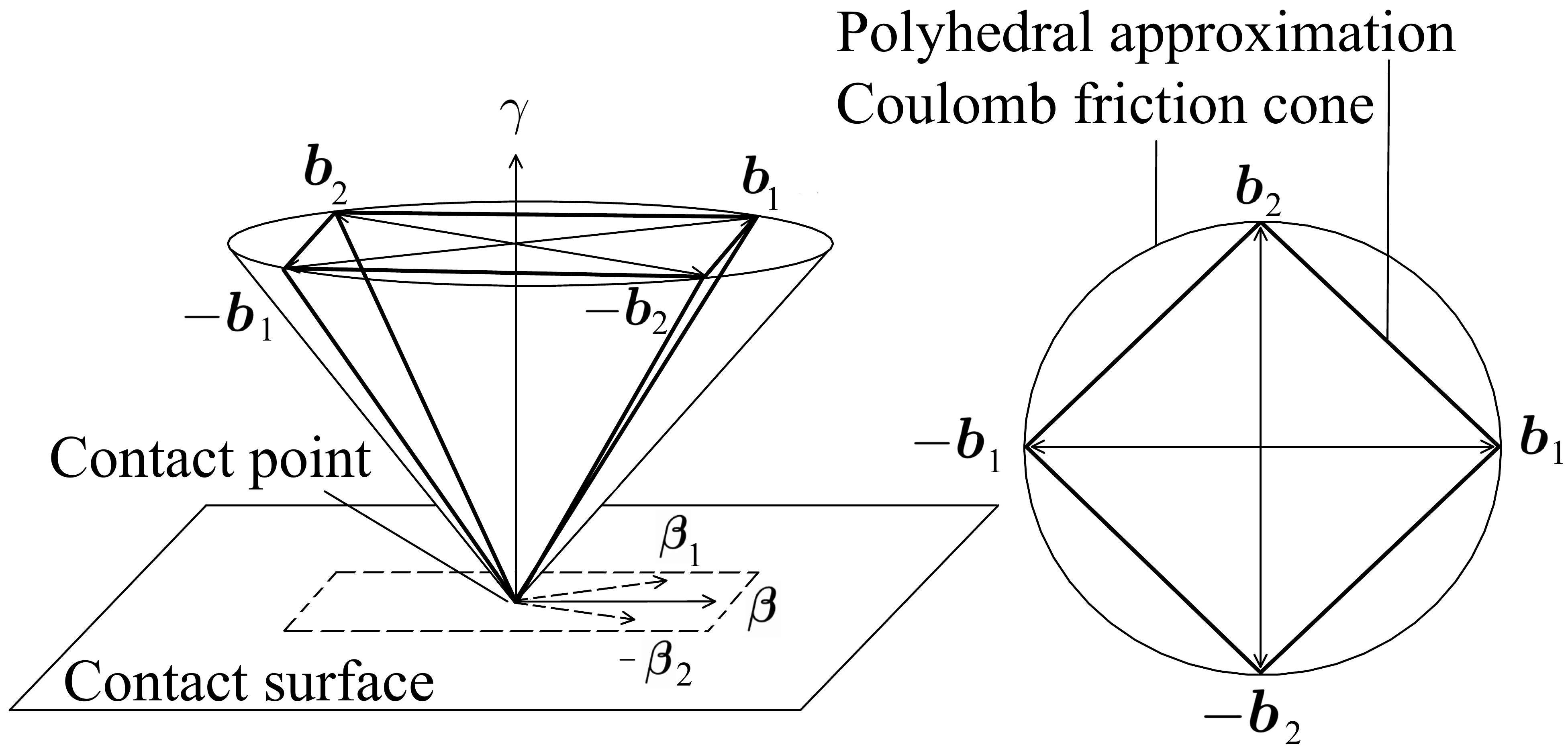}}
	\caption{The polyhedral (linear) approximation of the nonlinear friction cone (left). The basis vectors for the linearized cone (right).}\label{fig:cone_approx}
\end{figure}

Friction forces arise when two surfaces are in contact. While there exists a large variety of friction models, see \cite{popov_contact_2010} for an overview, Coulomb's simple dry friction law has proven very effective for robotics applications \cite{manchester_contact-implicit_2019,posa_direct_2014}.
In this friction model, the tangential friction force $\bs{\beta}$ is perpendicular to the contact normal force $\gamma$ and it is constrained to lie within the friction cone
\begin{equation}
	\lVert \bs{\beta} \rVert_2 \leq \gamma c_\mathrm{f}, \label{eq:cone_constraint_nonlinear}
\end{equation}
where $c_\mathrm{f}$ denotes the friction coefficient. 

The nonlinearity of \eqref{eq:cone_constraint_nonlinear}, especially for small normal forces, leads to numerically issues in computations \cite{nocedal_numerical_2006,steward_implicit_1996}. 
Therefore, a polyhedral approximation of the cone is often used, yielding a linearized description of \eqref{eq:cone_constraint_nonlinear} with friction forces $\bs{\beta}$:
\begin{subequations}\label{eq:cone_constraint_linear}
	\begin{align}
		\lVert \bs{\beta} \rVert_1 = \pmb{1}\hT\bs{\beta} &\leq \gamma c_\mathrm{f},\\
		\bs{\beta} &\geq \bs{0},
	\end{align}
\end{subequations}
where $\pmb{1} = [1 ~ \cdots ~ 1]\hT$. The comparison of nonlinear and linear friction cone is visualized in Fig. \ref{fig:cone_approx} and explained in detail in \cite{steward_implicit_1996}. 

The direction and magnitude of the friction forces can be derived with the maximum dissipation principle \cite{preclik_maximum_2018}.
This principle states that friction forces $\bs{\beta}$ maximize a body's kinetic energy dissipation rate $\frac{\mathrm{d}}{\mathrm{d}t}\mathcal{T}$:
\begin{subequations}\label{eq:max_dissip}
	\begin{alignat}{2}
		&\min_{\bs{\beta}} ~ &&\frac{\mathrm{d}}{\mathrm{d}t}\mathcal{T}(\bs{z}) = \min_{\bs{\beta}} ~ \dot{\bs{z}}\hT B(\bs{z})\hT \bs{\beta},\\
		&\text{s.t.}~&&\pmb{1}\hT\bs{\beta} \leq \gamma c_\mathrm{f},\\
		& &&\bs{\beta}\geq \bs{0},
	\end{alignat}
\end{subequations}
where $B = [\bs{b}_1 ~ -\bs{b}_1 ~ \cdots ~ \bs{b}_n ~ -\bs{b}_n]$ consists of the $2n$ basis vectors of the approximated friction cone visualized in Fig. \ref{fig:cone_approx} and maps the friction forces $\bs{\beta}$ into the dynamics.

The constraint \eqref{eq:contact_constraint} and optimization problem \eqref{eq:max_dissip} must be satisfied when calculating the dynamics of an articulated mechanism with environmental contact.

\subsection{Maximal Coordinates}
A detailed derivation of variational integrators in maximal coordinates without contact and friction can be found in \cite{brudigam_linear-time_2021}, and we will briefly summarize the key concepts.

In maximal coordinates, rigid bodies are described by their full configuration 
\begin{equation}\label{eq:zk}
	\bs{z} = \begin{bmatrix}
		\bs{x}\\
		\bs{q}
	\end{bmatrix},
\end{equation}
where $\bs{x} \in \mathbb{R}^3$ and $\bs{q} \in \mathbb{R}^4$ are the position and orientation (unit quaternion) in the global frame, respectively. Connections (joints) between bodies are formed with equality constraints
\begin{equation}
	\bs{g}(\bs{z}) = \bs{0}.
\end{equation}

The resulting equations of motion can be found by minimizing the action integral
\begin{equation}\label{eq:action}
	S = \int^{t_N}_{t_0} \mathcal{L}\left(\bs{z}(t),\dot{\bs{z}}(t)\right)~\mathrm{d}t + \int^{t_N}_{t_0} \bs{\lambda}\hT\bs{g}(\bs{z})~\mathrm{d}t,
\end{equation}
where $\mathcal{L}$ is the Lagrangian of the system and $\bs{\lambda}$ is a Lagrange multiplier (constraint force) enforcing the adherence to constraints $\bs{g}(\bs{z})$.

For a first-order variational integrator with time step $\Delta t$, the velocities are discretized as follows:
\begin{equation}\label{eq:dotzk}
	\dot{\bs{z}}_k(\bs{z}_k, \bs{z}_{k+1}) = \begin{bmatrix}
		\bs{v}_k\\
		\bs{\omega}_k
	\end{bmatrix},
\end{equation}
where
\begin{subequations}\label{eq:discvel}
	\begin{align}
		\bs{v}_k &= \frac{\bs{x}_{k+1}-\bs{x}_k}{\Delta t},\\
		\bs{\omega}_k &= \frac{2\bs{q}_{k}^{-1}\bs{q}_{k+1}}{\Delta t}.
	\end{align}
\end{subequations}
For details on the angular velocity, see \cite{brudigam_linear-time_2021}.

A (first-order) variational integrator can be obtained by discretizing \eqref{eq:action} over three time steps:
\begin{equation}\label{eq:discrete_action}
	S_\text{d} = \sum^{2}_{k=0} \left(\mathcal{L}(\bs{z}_k,\dot{\bs{z}}_k) + \bs{\lambda}_k\hT\bs{g}(\bs{z}_k)\right)\Delta t.
\end{equation}
Least action, i.e., minimizing \eqref{eq:discrete_action}, yields the implicit discretized equations of motion:
\begin{equation}\label{eq:dynamics}
	\nabla_{\bs{z}_1} S_\text{d} = -\bs{d} = \bs{0}.
\end{equation} 

The resulting nonlinear equations take the form
\begin{subequations}\label{eq:dynamics2}
	\begin{align}
		\bs{d}(\dot{\bs{z}}_{k+1},\bs{\lambda}_{k+1}) &= \bs{d}_0(\dot{\bs{z}}_{k+1}) - G(\bs{z}_{k+1})\hT\bs{\lambda}_{k+1} = \bs{0},\label{eq:implizit_constrained_dyn}\\
		\bs{g}(\bs{z}_{k+2}) &= \bs{0},\label{eq:implizit_constrained_dynb}
	\end{align}
\end{subequations}
where $\bs{d}_0$ are the unconstrained dynamics, and $G$ is the Jacobian of constraints $\bs{g}$ with respect to $\bs{z}$,
\begin{equation}
	G = \frac{\partial \bs{g}}{\partial \bs{z}},
\end{equation}
mapping the constraint forces into the dynamics. Note, that the constraints must be enforced for $\bs{z}_{k+2}$.

In order to simulate a system forward in time with this variational integrator given $\bs{z}_k$ and $\dot{\bs{z}}_k$, the next configuration $\bs{z}_{k+1}$ is trivially calculated with \eqref{eq:discvel}, and $\dot{\bs{z}}_{k+1}$ can subsequently be found by solving \eqref{eq:dynamics2}, for example with Newton's method. This integration scheme is the symplectic Euler method. Since the constraints are enforced at the position level, no constraint drift occurs, and therefore, no stabilization schemes are required.

\section{Linear-time Variational Integrator for Contact Dynamics}\label{sec:method}
In order to efficiently solve rigid body dynamics with contact and friction, we will first derive the dynamics in maximal coordinates, show the linear-time complexity in the number of bodies and constraints, and subsequently formulate an algorithm for solving the constrained dynamics.

\subsection{Variational Contact and Friction Dynamics}
In minimal coordinates, the signed distance function $\phi_{\text{min}}(\bs{z})$ (cf. \eqref{eq:contact_constraint}) and friction force mapping $B_{\text{min}}(\bs{z})$ (cf. \eqref{eq:max_dissip}) for a single contact point generally depend on the full kinematic chain. A change of configuration of any predecessor body of the contact point changes the contact point's location.

In contrast, in maximal coordinates, $\phi(\bs{z})$ and $B(\bs{z})$ only depend on the configuration of the respective body in contact.
Figure \ref{fig:contact_comp} visualizes these two perspectives. Because of the modular structure of maximal coordinates, each contact point only depends on its respective body, and therefore is (algebraically) independent of any other body or constraint, which enables efficient computations. 
In addition, all contact constraints have the same simple structure which allows for the analytical calculation of derivatives instead of costly online differentiation.

\subsubsection{Contact}
In maximal coordinates and discrete time, the signed distance function for a single contact point on the surface of a body, for example the foot in Fig. \ref{fig:contact_comp}, always takes on the form
\begin{equation}
	\phi(\bs{z}_k) = \bs{x}_k + R(\bs{q}_k)\bs{p} \geq 0, \label{eq:max_contact_constraint}
\end{equation}
where $\bs{x}_k$ is the position of the center of mass of the body, $\bs{p}$ is a vector from the body's center of mass to its surface in the local frame $\bs{q}_k$, and $R(\bs{q}_k)$ rotates $\bs{p}$ into the global frame. This rotation can also be performed directly with quaternion operations instead of constructing a rotation matrix.

In combination with the dynamics \eqref{eq:dynamics}, \eqref{eq:max_contact_constraint} is a nonlinear complementarity problem (NCP) with a Lagrange multiplier $\gamma_k$ acting as the contact normal force:
\begin{subequations}\label{eq:contact_comp}
	\begin{align}
		\phi(\bs{z}_k) &\geq 0,\\
		\gamma_k &\geq 0,\\
		\phi(\bs{z}_k)\gamma_k &= 0,
	\end{align}
\end{subequations}
for which we use the standard shorthand notation
\begin{equation}
	0\leq\phi(\bs{z}_k)\perp\gamma_k\geq0.
\end{equation}

\subsubsection{Friction}
We discretize the maximum dissipation principle with the discrete configuration \eqref{eq:zk} and velocity \eqref{eq:dotzk}:
\begin{subequations}\label{eq:max_dissip_max}
	\begin{alignat}{2}
		&\min_{\bs{\beta}_k} ~ &&\bs{v}_k\hT B_{\bs{x}}\hT \bs{\beta}_k + \bs{\omega}_k\hT B_{\bs{q}}(\bs{q}_k)\hT\bs{\beta}_k,\\
		&\text{s.t.} ~ &&\pmb{1}\hT\bs{\beta}_k \leq \gamma_k c_\mathrm{f},\\
		& &&\bs{\beta}_k\geq \bs{0}.
	\end{alignat}
\end{subequations}
In maximal coordinates, the mapping matrices are
\begin{align}
	B_{\bs{x}}\hT &= \begin{bmatrix}
		\bs{b}_1 ~ -\bs{b}_1 ~ \cdots ~ \bs{b}_n ~ -\bs{b}_n
	\end{bmatrix},\\
	B_{\bs{q}}(\bs{q}_k)\hT &= p^\times R(\bs{q}_k)\hT B_{\bs{x}}\hT,
\end{align}
where $p^\times$ is the skew-symmetric matrix formed from $\bs{p}$. Physically, the mapping of friction forces $\bs{\beta}_k$ at the contact point $\bs{p}$ with $B_{\bs{x}}$ generates a translational force, whereas the mapping with $B_{\bs{q}}$ generates a torque on the body.

As with the contact constraint, we formulate \eqref{eq:max_dissip_max} as an NCP with Lagrange multipliers $\psi_k$---the body's tangential velocity at the contact point---and $\bs{\eta}_k$:
\begin{subequations}\label{eq:fric_comp}
	\begin{align}
		B_{\bs{x}}\bs{v}_k + B_{\bs{q}}(\bs{q}_k)\bs{\omega}_k + \pmb{1}\psi_k - \bs{\eta}_k &= \bs{0},\\
		0 \leq c_\mathrm{f}\gamma_k - \pmb{1}\hT\bs{\beta}_k \perp \psi_k &\geq 0,\\
		\bs{0} \leq \bs{\beta}_k \perp \bs{\eta}_k &\geq \bs{0}.
	\end{align}
\end{subequations}

\begin{figure*}[t!] 
	\centering
	\resizebox{0.99\textwidth}{!}{\includegraphics{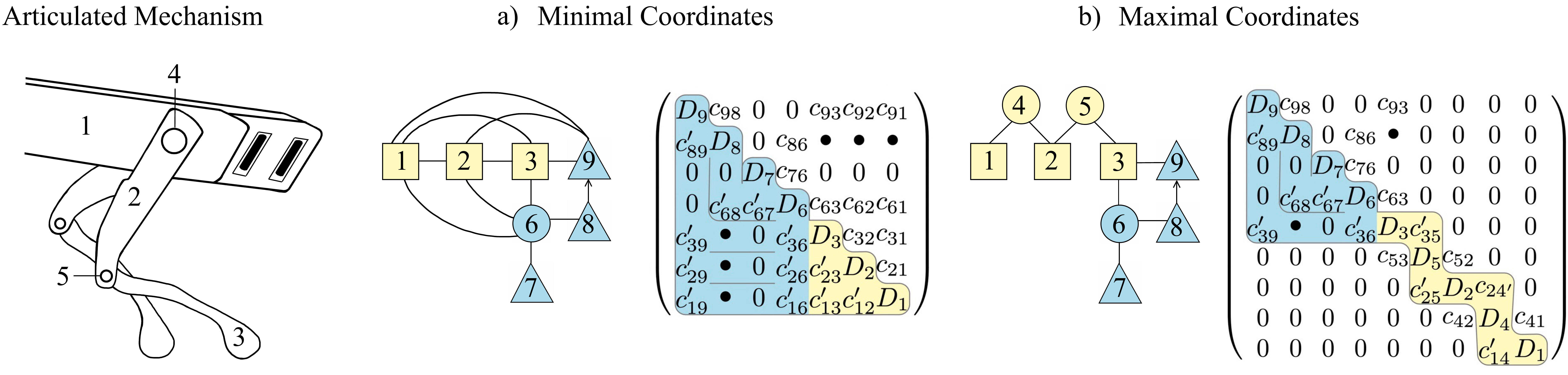}}
	\caption{Graphs and matrices for part of a quadrupedal mechanism with two joints and three links (left). In the graphs, links are drawn as squares, equality constraints as circles, and inequality constraints as triangles. Fill-ins in the matrices indicated as "$\bullet$". a) Graph and DFS-ordered matrix for minimal coordinates. b) Graph and DFS-ordered matrix for maximal coordinates. }\label{fig:structure_comp}
\end{figure*}

\subsubsection{Full dynamics}
The resulting contact normal force $\gamma_k$ and friction force $\bs{\beta}_k$ must be incorporated into the dynamics with the discretized least action principle \eqref{eq:discrete_action} in the same way as the constraint forces in \eqref{eq:implizit_constrained_dyn}, resulting in: 
\begin{equation}\label{eq:system_dyn}
	\begin{alignedat}{1}
		&\bs{d}_0(\dot{\bs{z}}_{k+1}) - G(\bs{z}_{k+1})\hT\lambda_{k+1}\\
		&~~~~~~~~~ - N(\bs{z}_{k+1})\hT\gamma_{k+1} - B(\bs{z}_{k+1})\hT\bs{\beta}_{k+1} = \bs{0},
	\end{alignedat}
\end{equation}
where 
\begin{equation}
	N = \frac{\partial \phi}{\partial\bs{z}},
\end{equation}
and
\begin{equation}
	B = [B_{\bs{x}} ~ B_{\bs{q}}].
\end{equation}

As a result, the complete NCP including the dynamics and equality constraints \eqref{eq:dynamics2}, as well as contact constraints \eqref{eq:contact_comp} and friction constraints \eqref{eq:fric_comp} can be stated as follows:
\begin{subequations}\label{eq:full_system}
	\begin{align}
		\bs{d}(\dot{\bs{z}}_{k+1},\bs{\lambda}_{k+1},\gamma_{k+1}, \bs{\beta}_{k+1}) &= \bs{0},\label{eq:full_sys_dyn}\\
		\bs{g}(\bs{z}_{k+2}) &=\bs{0},\label{eq:full_sys_eqc}\\
		B(\bs{z}_{k+1})\dot{\bs{z}}_{k+1} + \pmb{1}\psi_{k+1} - \bs{\eta}_{k+1} &= \bs{0},\label{eq:full_sys_fric}\\
		0\leq(c_\mathrm{f}\gamma_{k+1} - \pmb{1}\hT\bs{\beta}_{k+1})\perp\psi_{k+1}&\geq0,\label{eq:full_sys_psi}\\
		\bs{0}\leq\bs{\beta}_{k+1}\perp\bs{\eta}_{k+1}&\geq \bs{0},\label{eq:full_sys_eta}\\
		0\leq\phi(\bs{z}_{k+2})\perp\gamma_{k+1}&\geq0.\label{eq:full_sys_gamma}
\end{align}
\end{subequations}

For multiple bodies and constraints, all equations are stacked. External forces could also be included, and their integration is explained in \cite{brudigam_linear-time_2021}.

\subsection{Linear-Time Complexity}
If \eqref{eq:full_system} is solved with a matrix-based numerical approach, such as Newton's method, we can show linear-time complexity both in the number of bodies and number of contact points of a system. 

Solving \eqref{eq:full_system} with any variant of Newton's method requires at each iteration to solve a linear system of equations
\begin{equation}\label{eq:linsys}
	F\Delta\bs{a} = \bs{r},
\end{equation}
where $\bs{r}$ are the active equations of \eqref{eq:full_system}, $F$ is the Jacobian of the active equations of \eqref{eq:full_system}, and $\Delta\bs{a}$ is the search direction to be found. 

Figure \ref{fig:structure_comp} shows exemplary matrices $F$ and associated graphs for the front right part of a quadrupedal robot in minimal and maximal coordinates. Nodes 1, 2, and 3 are the dynamics equations \eqref{eq:full_sys_dyn} associated with the trunk, upper leg, and lower leg, respectively. Nodes 4 and 5 are the equality constraints \eqref{eq:full_sys_eqc} for the joints in maximal coordinates. Node 6 represents the friction constraint \eqref{eq:full_sys_fric} for the foot, with nodes 7 and 8 being the limits on the total friction force \eqref{eq:full_sys_psi} and $\bs{\beta}$ \eqref{eq:full_sys_eta}, respectively. Node 9 is the contact inequality constraint \eqref{eq:full_sys_gamma}. 

Factorization and back-substitution (i.e., Gaussian elimination) is used to solve \eqref{eq:linsys}, and we exploit the sparsity structure in this system to minimize the number of operations needed. Performing a depth-first search (DFS) along the graph of \eqref{eq:linsys} gives us a processing order for the factorization that minimizes the number of fill-ins---zeros changing to non-zeros during factorization---and therefore minimizes the number of required operations. The matrices in Fig. \ref{fig:structure_comp} have been reordered according to a depth-first search, with the first found node placed at the bottom right.

For a matrix with an associated acyclic graph, no fill-ins will be created for a reordered matrix, and the factorization and back-substitution can be performed with linear-time complexity \cite{duff_direct_2017}. Cycles in a graph create fill-ins for nodes that are part of the cycle.

As can be seen in Fig. \ref{fig:structure_comp}, the contact and friction constraints form a cycle with the associated $n$ bodies in minimal coordinates and 1 body in maximal coordinates. These structures result in a fixed number of fill-ins for each set of contact constraints. Therefore, for $c$ contact points, we have $c$ sets of contact constraints that can be solved independently with a fixed number of operations per set, resulting in $O(c)$ complexity in regards to the number of contact points. This argument holds for both minimal and maximal coordinates.

Since in maximal coordinates, contact constraints are only connected to a single body, adding $n$ additional bodies and joint constraints does not effect the processing of the contact constraints and therefore we still obtain $O(c)$ complexity for the contact constraints. If the rest of the system is acyclic, it can be processed with $O(n)$ complexity \cite{brudigam_linear-time_2021}, and we achieve a total complexity of $O(n+c)$. 

In contrast, in minimal coordinates, each set of contact constraints generally forms a cycle with every body of the system, and so we obtain $O(cn)$ additional fill-ins, preventing linear-time complexity.

Note, that contact between bodies of a mechanism could also be treated with such a matrix-based method, but such contacts would create cycles and the linear-time property no longer directly applies.

\subsection{Interior-Point Method}
We solve \eqref{eq:full_system} with an interior-point method (see \cite{nocedal_numerical_2006} for details) to reduce the numerical difficulties of the contact constraints. However, the graph-based complexity argument holds for any matrix-based solver.

For the interior-point method, slack variables $\bs{s}$ are introduced for the inequality constraints and the complementarity constraints are initially relaxed by $\mu$ (time-step indices are dropped for clarity): 
\begin{subequations}\label{eq:ip_system}
	\begin{align}
		\bs{d}(\bs{z},\bs{\lambda},\gamma, \bs{\beta}) &= \bs{0},\\
		\bs{g}(\bs{z}) &=\bs{0},\\
		B(\bs{z})\dot{\bs{z}} + \pmb{1}\psi - \bs{\eta} &= \bs{0},\\
		(c_\mathrm{f}\gamma - \pmb{1}\hT\bs{\beta}) - s_\psi &= 0,\\
		\bs{\beta} - \bs{s}_{\bs{\eta}} &= \bs{0},\\
		\phi(\bs{z}) - s_\gamma &= 0,\\
		0\leq s_\psi\perp^{\mu}\psi&\geq 0,\\
		\bs{0}\leq\bs{s}_{\bs{\eta}}\perp^{\mu}\bs{\eta}&\geq \bs{0},\\
		0\leq s_\gamma\perp^{\mu}\gamma&\geq 0,
	\end{align}
\end{subequations}
where $\bs{a}\perp^{\mu}\bs{b}$ indicates $\bs{a}\hT\bs{b}=\mu$ compared to $\bs{a}\perp\bs{b}$ indicating $\bs{a}\hT\bs{b}=0$. 

Subsequently, Newton's method is applied on the modified equations \eqref{eq:ip_system}, where a line search ensures feasibility of all slack variables and multipliers. Additionally, $\mu$ is initially set to any non-zero positive value, but it is reduced to zero during the Newton iterations. In this way, the original complementarity problem is recovered and we preserve rigid contacts.

\section{Experiments and Simulations}\label{sec:eval}
We evaluate our algorithm by demonstrating the linear-time complexity in test scenarios. Additionally, we provide two simulation examples of robotic systems to show the applicability of the proposed method to real systems. 

The algorithm was implemented in the programming language Julia \cite{bezanson_julia_2017}. The code for our algorithm and all experiments is available at \url{https://github.com/janbruedigam/ConstrainedDynamics.jl/}. All experiments were performed on an ASUS ZenBook with an Intel i7 processor.

\subsection{Linear-Time Complexity}
As described in Sec. \ref{sec:method}, in maximal coordinates we can achieve linear-time complexity both in regards to the number of contact constraints and the number of bodies in a mechanism. We show $O(c)$ complexity for a single rigid body with $c$ contact points, and $O(c+n)$ complexity for a system consisting of $n$ bodies and $c=n+1$ contact points. 

All simulations ran with a time step of $\Delta t=0.01$ms. The termination condition for Newton's method at each time step was a converged solution $\lVert\bs{r}\rVert<\epsilon$, where $\epsilon=10^{-6}$. For each simulation, the best
timing result of 100 runs was taken to eliminate right-skewing computer noise. For both experiments, we simulated 100 time steps for each run. The friction coefficient was $c_\mathrm{f}=0.2$.

\subsubsection{Single Body}
The single rigid body is a cylinder with mass $m=1$kg, height $h=0.1$m, and radius $r = 0.5$m. We define an even number of contact points equally spaced on the circular base, ranging from 2 contact points to 100 contact points. 

\begin{figure}[t] 
	\centering
	\resizebox{0.45\textwidth}{!}{\includegraphics{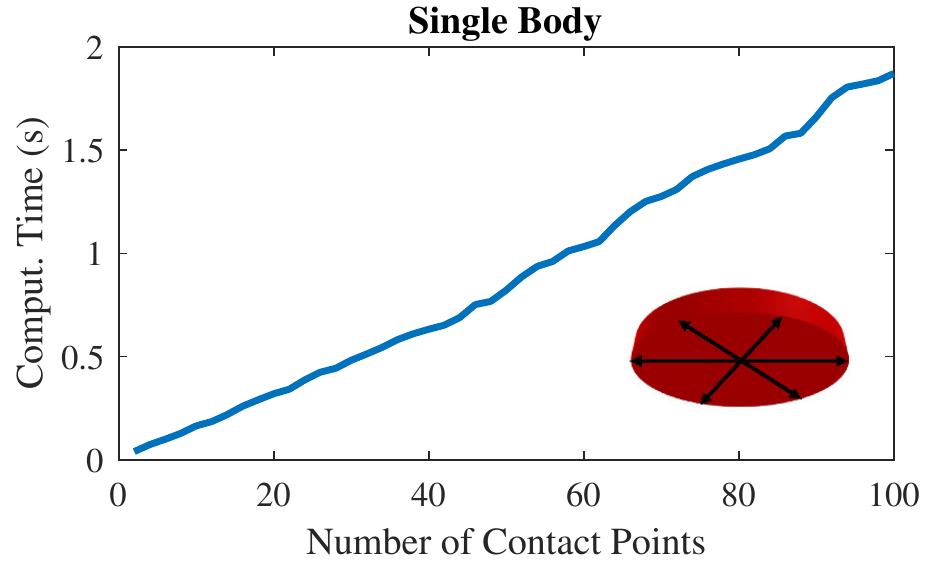}}
	\caption{Linear scaling behavior for simulating a single body (bottom right) with $c$ contact points.}\label{fig:disc}
\end{figure}

Figure \ref{fig:disc} shows the resulting computation time for the cylinder with $c$ contact points. The $O(c)$ computational complexity is clearly visible.

\subsubsection{Multiple Bodies}
As a system with multiple bodies, we chose a chain of $n$ links connected by spherical joints, resembling, for example, a snake robot. The cylindrical links have a mass $m=1$kg, length $l=1$m, and radius $r=0.05$m. Each link has a single contact point at its lower end, and the first link has an additional contact point at its upper end, resulting in $c=n+1$ contact points. We evaluated 2 to 100 links.

\begin{figure}[t] 
	\centering
	\resizebox{0.45\textwidth}{!}{\includegraphics{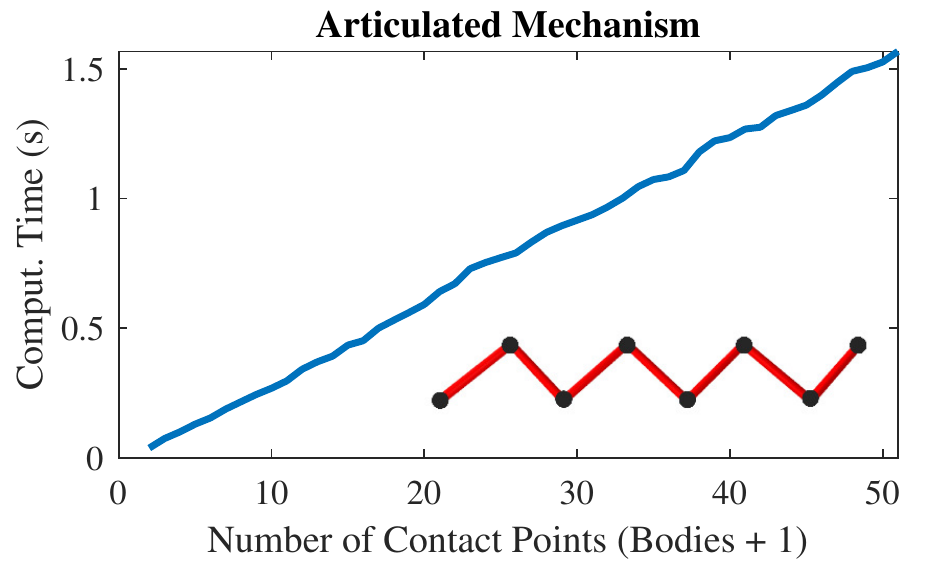}}
	\caption{Linear scaling behavior for simulating an articulated mechanism (bottom right) with $n$ bodies and $c=n+1$ contact points.}\label{fig:snake}
\end{figure}

The results for the chain of $n$ links with $c=n+1$ contact points is displayed in Fig. \ref{fig:snake}. As before, the $O(n+c)$ computational complexity can be clearly seen.

\subsection{Simulation Examples}
As simulation examples resembling real robotic systems, we chose a three-dimensional hopper and a three-dimensional quadruped. 

A time step of $\Delta t=0.001$ms was chosen. As before, the termination condition for each time step was a converged solution $\lVert\bs{r}\rVert<\epsilon$, where $\epsilon=10^{-6}$. 

\subsubsection{3D Hopper}
The hopper consists of a floating-base center sphere with mass $m=1$kg and radius $r=0.1$m, and a cylindrical pole with mass $m=0.2$kg, length $l=0.8$m, and radius $r=0.05$m, that is attached to the center sphere with an actuated prismatic joint. The center sphere of the hopper can apply a torque in all three dimensions, which could be achieved with flywheels as in satellites. The pole has a contact point at its lower end with a friction coefficient of $c_\mathrm{f}=0.2$. 

For the simulation, the hopper performs a single jump and does a full turn around the $[1 ~ 1 ~ 0]\hT$ axis while in the air. This maneuver demonstrates the contact and friction dynamics in combination with a quaternion-based orientation description for singularity-free movements. 

\begin{figure}[t] 
	\centering
	\resizebox{0.45\textwidth}{!}{\includegraphics{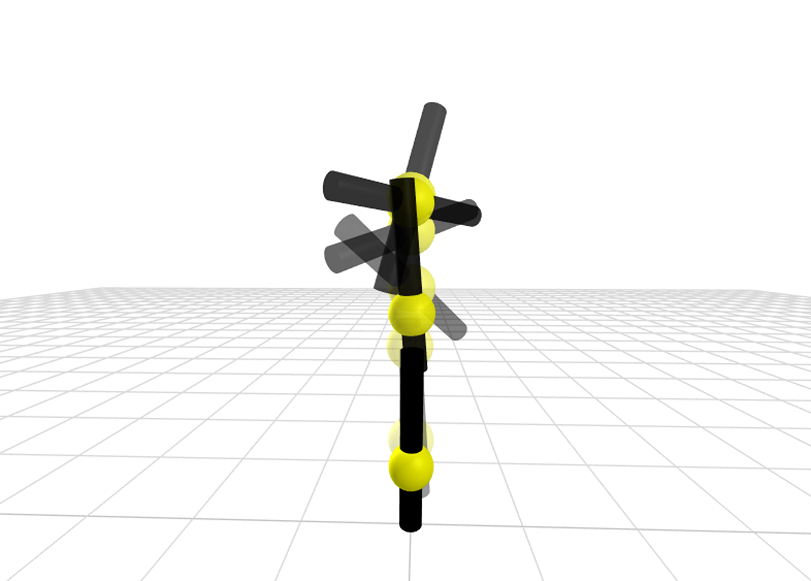}}
	\caption{Visualization of the simulation of the 3D hopper robot performing a full turn in the air.}\label{fig:hopper}
\end{figure}

The simulated trajectory of the hopper is visualized in Fig. \ref{fig:hopper}. The minimum signed distance of the pole's contact point was $0.0$m, i.e., the pole being initially on the ground.

\subsubsection{Quadruped}
Walking robots, such as bipeds or quadrupeds, have recently found increasing interest for real-world applications, for example \textit{Boston Dynamics' Spot} robot. For our evaluation, we simulate the \textit{UnitreeRobotics} quadruped consisting of a floating-base trunk and four legs with three segments each (shoulder, upper leg, lower leg). For the simulation, the quadruped performs a simple periodic walking gait.

\begin{figure}[t] 
	\centering
	\resizebox{0.45\textwidth}{!}{\includegraphics{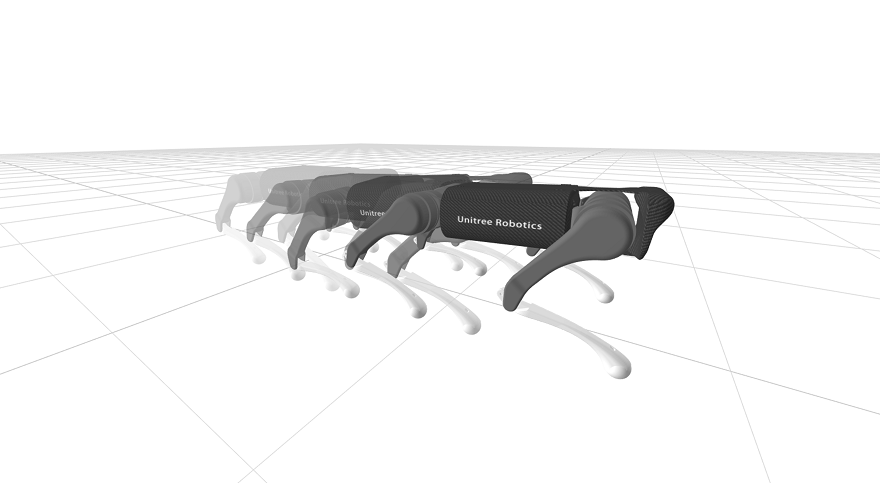}}
	\caption{Visualization of the simulation of the 3D quadrupedal robot walking with a periodic gait.}\label{fig:quadruped}
\end{figure}

\begin{figure}[t] 
	\centering
	\resizebox{0.45\textwidth}{!}{\includegraphics{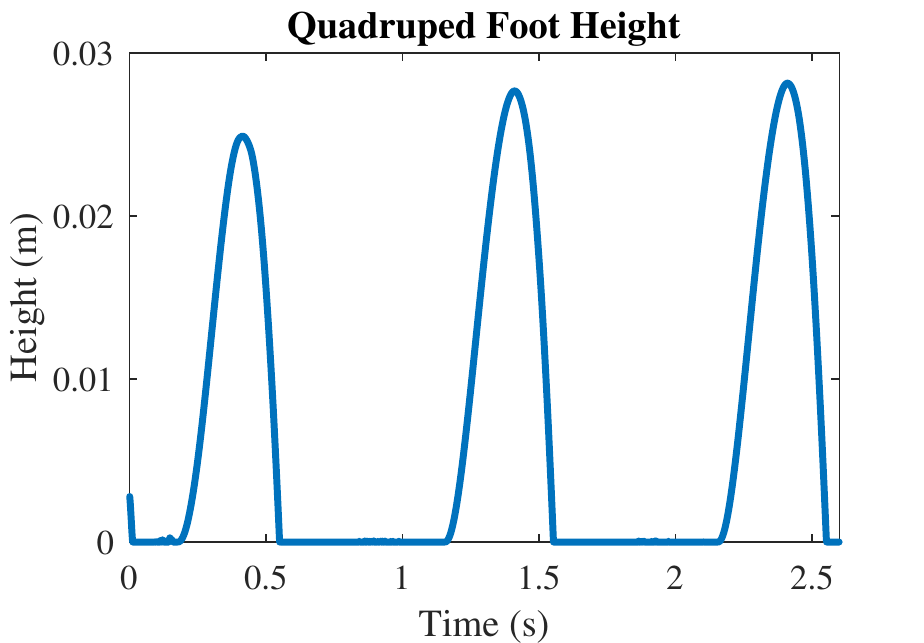}}
	\caption{The signed distance function over time for the front right foot of the quadruped. No penetration of the ground occurred.}\label{fig:sdf_foot}
\end{figure}

The simulated trajectory of the quadruped is visualized in Fig. \ref{fig:quadruped}, demonstrating the applicability of our algorithm to the simulation of real-life robots. An exemplary signed distance function for one of the feet is given in Fig. \ref{fig:sdf_foot}. The minimum signed distance of the contact points of all feet was $1.5\cdot 10^{-9}$m.

\section{Conclusions}\label{sec:conclusions}
We have proposed an algorithm for simulating rigid-body dynamics encountering environmental contact and friction with linear-time complexity both in the number of contact points and the number of bodies. The constrained dynamics equations are discretized with a variational integrator and solved with an interior-point method to reduce numerical difficulties while maintaining physical accuracy and preventing constraint drift. The evaluation of the implemented algorithm confirmed the theoretical algorithm properties and demonstrated the applicability to robotic systems.

In addition to general simulation tasks, our proposed method could be used to improve the computational efficiency of the contact-implicit model-predictive control formulation in \cite{cleach_linear_2021}. 

An extension of our proposed method to treating contact between multiple bodies would be needed for treating grasping tasks. Due to the intra-mechanism contact, a modification to the proposed method is required to achieve linear-time complexity. Additionally, an investigation whether the improved control performance demonstrated with maximal coordinates \cite{brudigam_linear-quadratic_2021,knemeyer_minor_2020} also extends to systems with contact interactions could be beneficial.

\section*{Acknowledgements}
This work was supported by the European Commission grant H2020-ICT-871295 (``SeaClear'': SEarch, identificAtion and Collection of marine Litter with Autonomous Robots).
\IEEEtriggeratref{13}
\bibliography{bib}
\bibliographystyle{ieeetr}

\end{document}